\def\year{2020}\relax
\documentclass[letterpaper]{article} % DO NOT CHANGE THIS
\usepackage{aaai20}  % DO NOT CHANGE THIS
\usepackage{times}  % DO NOT CHANGE THIS
\usepackage{helvet} % DO NOT CHANGE THIS
\usepackage{courier}  % DO NOT CHANGE THIS
\usepackage[hyphens]{url}  % DO NOT CHANGE THIS
\usepackage{graphicx} % DO NOT CHANGE THIS
\usepackage{graphicx}  % DO NOT CHANGE THIS
\usepackage{amsmath}
\DeclareMathOperator*{\argmax}{argmax}

\usepackage{subcaption}
\usepackage{multirow}
\title{Likelihood Ratios and Generative Classifiers for Unsupervised Out-of-Domain Detection In Task Oriented Dialog}
\author{ \Large \textbf{Varun Gangal\textsuperscript{\rm 1, \rm 2 \thanks{Work done by author while interning at Facebook Conv AI}},  Abhinav Arora\textsuperscript{\rm 2},  Arash Einolghozati\textsuperscript{\rm 2},  Sonal Gupta\textsuperscript{\rm 2}}\\ % All authors must be in the same font size and format. Use \Large and \textbf to achieve this result when breaking a line
\textsuperscript{\rm 1} Language Technologies Institute, Carnegie Mellon University, Pittsburgh, PA 15213 \\
\textsuperscript{\rm 2} Facebook Conversational AI, Menlo Park, CA 94303\\
{vgangal@andrew.cmu.edu} {\{abhinavarora,arashe,sonalg\}}@fb.com % email address must be in roman text type, not monospace or sans serif
}

\begin{document}

\maketitle

\begin{abstract}
The task of identifying out-of-domain (\textit{OOD}) input examples directly at test-time has seen renewed interest recently due to increased real world deployment of models. In this work, we focus on OOD detection for natural language sentence inputs to task-based dialog systems. Our findings are three-fold:

First, we curate and release \textsc{ROSTD} (\textbf{R}eal \textbf{O}ut-of\\-Domain \textbf{S}entences From \textbf{T}ask-oriented \textbf{D}ialog) - a dataset of $4K$  \textit{OOD} examples for the publicly available dataset from \cite{schuster2018cross}. In contrast to existing settings which synthesize \textit{OOD} examples by holding out a subset of classes, our examples were authored by annotators with apriori instructions to be out-of-domain with respect to the sentences in an existing dataset.

Second, we explore likelihood ratio based approaches as an alternative to currently prevalent paradigms. Specifically, we reformulate and apply these approaches to natural language inputs. We find that they match or outperform the latter on all datasets, with larger improvements on non-artificial \textit{OOD} benchmarks such as our dataset. Our ablations validate that specifically using likelihood ratios rather than plain likelihood is necessary to discriminate well between \textit{OOD} and in-domain data.

Third, we propose learning a generative classifier and computing a marginal likelihood (ratio) for \textit{OOD} detection. This allows us to use a principled likelihood while at the same time exploiting training-time labels. We find that this approach outperforms both simple likelihood  (ratio) based and other prior approaches. We are hitherto the first to investigate the use  of generative classifiers for \textit{OOD} detection at test-time. 
\end{abstract}

\section{Introduction}
\label{sec:intro}

With increased use of ML models in real life settings, it has become imperative for them to self-identify, at test-time, examples on which they are likely to fail due to them differing significantly from the model's training time distribution.  

In particular, for state-of-the-art deep classifiers such as those used in vision and language tasks, it has been observed that the raw probability value is over calibrated \cite{guo2017calibration} and can have high values even for \textit{OOD} inputs. This necessitates having an auxiliary mechanism to detect them. 

This task is not in entirety novel, and has historically been explored  in related forms under various names such as \textit{one class classification}, \textit{open classification} etc. The recent stream of work on this  started with \cite{hendrycks2016baseline}, which proposed benchmark datasets for doing this on vision problems.
\cite{liang2017enhancing} find that increasing the softmax temperature $\tau$ makes the resultant probability more discriminative for \textit{OOD} Detection. \cite{lee2018simple} propose using distances to per-class Gaussians in the intermediate representation learnt by the classifier. Specifically, a Gaussian is fit for each training class from all training points in that class . \cite{ren2019likelihood} show that ``correcting" likelihood with likelihood from a ``background" model trained on noisy inputs is better at discriminating out of distribution examples.  Recently, \cite{lin2019deep} propose using an old measure from the data mining literature named \textsc{LOF} \cite{breunig2000lof}  in the space of penultimate activations learnt by a classifier.

Apart from \cite{lin2019deep} and few others, a majority of the prior work uses vision problems and datasets, often image classification as the setting in which to perform \textit{OOD} Detection. Certain methods, such as input gradient reversal from \cite{liang2017enhancing} or an end-to-end differentiable Generative Adversarial Network (GAN) as in \cite{lee2017training} are not directly applicable for natural language inputs. Furthermore, image classification has available several benchmarks with a similar label space (digits and numbers) but differing input distributions, such as \textit{MNIST}, \textit{CIFAR} and \textit{SVHN}. Most of these works exploit this fact for their experimental setting by picking one of these datasets as \textit{ID} and the other as \textit{OOD}. 
In this work, we attempt to address these lacunae and specifically explore which \textit{OOD} detection approaches work well on natural language, in particular, intent classification. 

This problem is greatly relevant for task oriented dialog systems since intent classification can receive user intents which are sometimes not in any of the domains defined by the current ontology or downstream functions. In particular, \textbf{unsupervised} $OOD$ detection approaches are important as it is difficult to curate this kind of data for training because 
\begin{enumerate}
\item The size of in-domain data to train on can become arbitrarily large as the concerned dialog system gets more users and acquires the ability to handle newer intent classes. After a point, it becomes impractical to continue curating newer $OOD$ examples for training in proportion to the in-domain data. From there on, class imbalance would keep increasing.
\item  By definition, \textit{OOD} is an open class. For natural language intents, utterances can demonstrate diverse sentence phenomenena such as slang, rhetorical questions, code mixed language, etc. User data can exhibit a large range of \textit{OOD} behaviours, all of which may be difficult to encapsulate using a limited set of OOD examples at training time. 
\end{enumerate}

To the best of our knowledge, this is the first application of likelihood ratios approach for \textit{OOD} Detection in natural language. Overall, our contributions are as follows:
\begin{enumerate}
    \item We release \textsc{ROSTD}, a novel dataset\footnote{Our dataset is available at \url{github.com/vgtomahawk/LR_GC_OOD/blob/master/data/fbrelease/OODrelease.tsv}} of $4500$ \textit{OOD} sentences for intent classification. We observe that existing datasets for \textit{OOD} intent classification are 
    \begin{itemize}
        \item  too small ($<$1000 examples)
        \item create \textit{OOD} examples synthetically 
    \end{itemize}
    We show that performing \textit{OOD} detection on \textsc{ROSTD} is more challenging than the synthetic setting where \textit{OOD} examples are created by holding out some fraction of intent classes. We further describe this dataset in \S \ref{sec:data} .
    \item We show that using the marginal likelihood of a generative classifier provides a principled way of both incorporating the label information [like classifier uncertainty based approaches] while at the same time testing for \textit{ID} vs \textit{OOD} using a likelihood function.
    \item 
    We show that using likelihood with a correction term from a ``background" model, based on the formalism proposed in \cite{ren2019likelihood}, is a much more effective approach than using the plain likelihood. We propose multiple ways of training such a background model for natural language inputs.
    \item Our improvements hold on multiple datasets - both for our dataset as well as the existing \textsc{SNIPS} dataset \cite{coucke2018snips}.

\end{enumerate}

\section{Methods} \label{sec:methods}

\begin{figure*}[t!]
    \captionsetup{font=small}
    \centering
    \includegraphics[width=0.8\textwidth]{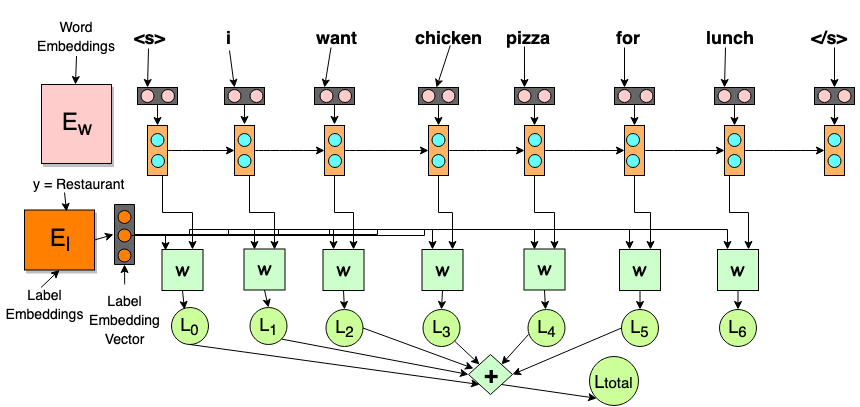}
    \caption{We illustrate the architecture of our generative classifier. $E_{w}$ and $E_{l}$ are the word embeddings and the label embeddings respectively. The hidden state is concatenated with the label embedding for \textit{``Restaurant"} before passing through the output layer $W$. Best seen in color.}
    \label{fig:modelDiagram}
    %\vspace{-0.4cm}
\end{figure*}

All our methods attempt to estimate a score which is indicative of the data point being \textit{OOD}. For an input $x$, we refer to this function as $\eta(x)$. This function may additionally be parametrized by the classifier distribution $\hat{P}$ or the set of nearest neighbors $N_k(x)$, based on the specific method in use. 

Some of our evaluation measures are threshold independent. In this case, $\eta(x)$ can be directly evaluated for its goodness of detecting OOD points. For measures which are threshold dependent, an optimal threshold which maximizes macro-F1 is picked using the values of $\eta(x)$ on a validation set.

\subsection{Maximum Softmax Probability}
\textit{Maximum Softmax Probability}, or \textsc{MSP} is a simple and intuitive baseline proposed by \cite{hendrycks2016baseline}. \textsc{MSP} uses the maximum probability 1-$\max_{y} \hat{P}(y|x)$ as $\eta(x,\hat{P})$. The lesser ``confident" the classifier $\hat{P}$ is about its predicted outcome i.e the argmax label, the greater is $\eta(x,\hat{P})$. Typically,
\begin{align*}
\hat{P}(y|x) &= \frac{e^{\frac{z_{y}}{\tau}}}{\sum_{y} e^{\frac{z_{y}}{\tau}} }    
\end{align*}
Here, $z_{y}$ denotes the logit for label $y$ while $\tau$ denotes the softmax temperature. Increasing $\tau$ smoothens the distribution while decreasing $\tau$ makes it peakier. We also try increased values of $\tau$ as they were shown to work better by \cite{liang2017enhancing}.

\subsection{Softmax Entropy}
Alternatively, both \cite{lee2017training} and \cite{hendrycks2018deep} propose using either of the following\footnote{They differ only by a constant and have the exact same minimas, as we show in Appendix 1.1. Appendix can be read at \url{github.com/vgtomahawk/LR_GC_OOD/tree/master/appendix}}:
\begin{enumerate}
\item Entropy $H_Y{\hat{P}(y|x)} = - \sum  _ {y \in Y} \hat{P}(y|x) \log \hat{P}(y|x) $ as $\eta(x,\hat{P})$ 
\item Negative KL Divergence $-KL(\hat{P}|\mathbf{U})$ w.r.t the uniform distribution over labels $\mathbf{U}$.
\end{enumerate}
We refer to this method as $-KL(\hat{P}|\mathbf{U})$ in our experiments\footnote{One distinction from the two cited papers is that they use $-KL(\hat{P}|\mathbf{U})$ merely as an auxiliary training objective, and end up using \textsc{MSP} as the $\eta$ at test time. In contrast, we  explicitly use $-KL(\hat{P}|\mathbf{U})$ as $\eta$}. Here, we also experiment with a variant of this method which replaces $\mathbf{U}$ with $\mathbf{R}$, where $\mathbf{R}(y)= \frac{\sum_{i=1}^{i=m} \mathbf{1}{(y_i =y)}}{|m|}$, or the fraction of class $y$ in the training set. We expect this variant to do better when \textit{ID} classes are not distributed equally.  

\subsection{Local Outlier Factor} \label{subsec:LOF}
\textsc{LOF}, proposed by \cite{breunig2000lof}, is a measure based on local density defined with respect to nearest neighbours. Recently, \cite{lin2019deep} effectively used LOF in the intermediate representation learnt by a classifier for \textit{OOD} detection of intents. The \textsc{LOF} measure can be defined in three steps:

\begin{enumerate}
    \item First, $reachdist_{k}(A,B) = max ( kdist(B), d(A,B))$ . Here, $kdist(B)$ is the distance to the kth nearest neighbor, while $d$ is the distance measure being used. Intuitively, $reachdist_k(A,B)$ is lower-bounded by $kdist(B) \forall A$, but can become arbitrarily large.
    \item Next, define a measure named local reachability density or \textit{lrd}. This is simply the reciprocal of the average $reachdist_{k}$ for $A$
    \begin{align*}
        lrd(A) = \frac{|N_{k}(A)|}{\sum_{B \in N_{k}(A) reachdist_{k}(A,B)}}
    \end{align*}
    \item Lastly, \textsc{LOF} is defined as
    \begin{align*}
    LOF_{k}(A) &= \frac{\sum_{B \in N_{k}(A)} \frac{lrd(B)}{lrd(A)}}{|N_k(A)|} 
    \end{align*}
\end{enumerate}
 Intuitively, if the ``density" around a point's nearest neighbours is higher than its own ``density", the point will have a higher \textsc{LOF}. Points with a higher $LOF$ score are more likely to be \textit{OOD}.

\cite{lin2019deep} further show that using the \textit{large margin cosine} loss or \textit{LMCL}, works better than the typical combination of softmax + cross entropy. 
\begin{align*}
\hat{P}(y|x) &= \frac{e^{\frac{{ {\overline{w_{y}}}}^{T}\overline{x}-m }{\tau}}}{\sum_{y} e^{\frac{  {\overline{w_{y}}}^{T}\overline{x} -m}{\tau}} }    
\end{align*} \label{eq:LMCL}

Here, $m$ denotes the margin. $w_{y}$ denotes the row in the final linear layer weight matrix corresponding to the label $y$. $x$ denotes the penultimate layer activations which are input to the final layer. We use $\overline{v}$ to denote the normalized $v$, i.e $\frac{v}{||v||}$.

We denote this approach as \textsc{LOF+LMCL}. We directly use the author's implementation \footnote{https://github.com/thuiar/DeepUnkID} for this approach. 

%\subsection{Papernot's Score}
%We refer to the measure proposed by \cite{papernot2018deep} as \textsc{PPS}. \textsc{PPS} first computes an uncertainty score based on the average uncertaintqy of $\hat{P}$ over $k$ nearest neighbors from the training set. Then, based on a ``calibration set"  , the fraction of points with greater uncertainty gives $\eta(x,\hat{P})$.

%\subsection{Mahalanobis Distance}

\subsection{Likelihood} \label{subsec:Lsimple}
Here, $\eta(x)$ is the likelihood $\mathbf{L}_{simple} = \hat{P}_{\mathbf{M}}(x)$ according to a  model $\mathbf{M}$ trained on the \textit{ID} training points. In the simplest case,  $\mathbf{M}$ is simply a left-to-right  language model learnt on all our training sentences. Later, we discuss another class of models which can give a valid likelihood, which we name $\mathbf{L}_{gen}$. 
 %\ref{subsec:Lgen} %In \S\S \ref{subsec:Lgen}
\subsection{Likelihood Ratio}
\cite{nalisnick2018deep,choi2018generative} found that likelihood is poor at separating out \textit{OOD} examples; in some cases even assigning them higher likelihood than the \textit{ID} test split. \cite{ren2019likelihood} make similar observations for detecting \textit{OOD} DNA sequences. They then propose the concept of \textit{likelihood ratio} or \textsc{LLR} based methods, which we briefly revisit here.

Let ${\hat{P}}_{\mathbf{M}}(x)$ and ${\hat{P}}_{\mathbf{B}}(x)$ denote the probability of $x$ according to the model $\mathbf{M}$ and a background model $\mathbf{B}$ respectively. $\mathbf{M}$ is trained on the training set ${\{x_i\}}_{i=1}^{i=m}$, while $\mathbf{B}$ is trained on noised samples from the training set.  Let ${x_{1}}^{i}$ denote the prefix $x_{1} x_{2} \ldots x_{i-1}$. The \textsc{LLR} is  derived as in Equation \ref{eq:LLR}
\begin{align*} 
\footnotesize
  LLR_{\mathbf{M},\mathbf{B}}(x)  &= \frac{{\hat{P}}_{M}(x)}{{\hat{P}}_{B}(x)} \\
  LLR_{\mathbf{M},\mathbf{B}}(x)  &= \frac{\Pi_{i=1}^{i=|S|} {\hat{P}}_{\mathbf{M}}(x_{i}|{x_{1}}^{i})}{\Pi_{i=1}^{i=|S|} {\hat{P}}_{\mathbf{B}}(x_{i}|{x_{1}}^{i})} \\
  \log {LLR_{\mathbf{M},\mathbf{B}}(x)} &= \sum_{i=1}^{i=|S|}  \log { {\hat{P}}_{\mathbf{M}}(x_{i}|{x^{i}_1})}  -  \log {{\hat{P}}_{\mathbf{B}}(x_{i}|{x_{1}^{i}})} 
\normalsize
\end{align*} \label{eq:LLR} % 
The intuition was that ``surface-level" features might be causing the \textit{OOD} points to be assigned a reasonable probability. The hypothesis is that the background model would capture these ``surface-level" features which also persist after noising and remove their influence on being divided out. If it were so, $LLR_{\mathbf{M},\mathbf{B}}(x)$ would be a better choice for $\eta(x)$.

\subsubsection{How to introduce noise?}
It is a common practice in vision to add noise or perturb images slightly by adding a Gaussian noise vector of small magnitude. Since natural language utterances are sequences of discrete words, this does not extend directly to them.  

A simple alternative to introduce noise into natural language inputs is by random word substitution. Word substitution based noise has a long precedent of use, from negative sampling as in \textit{word2vec} \cite{mikolov2013distributed,goldberg2014word2vec} to autoencoder-like objectives like \textsc{BERT} \cite{devlin2018bert}. 

More specifically, with probability $p_{noise}$, we substitute each word $w$ with word $w^{\prime}$ sampled from the distribution $\mathbf{N}(w^{\prime})$. $p_{noise}$ is a hyperparameter to be tuned. We experiment with 3 different choices of $\mathbf{N}$:
\begin{enumerate}
    \item \textsc{UNIFORM}: $\mathbf{N}(w^{\prime}) = \frac{1}{|W|}$ i.e each word $w^{\prime} \in W$ is equally likely.
    \item \textsc{UNIGRAM}: $\mathbf{N}(w^{\prime}) = \frac{f(w^{\prime})}{\sum_{w \in W} f(w)}$ i.e a word $w^{\prime}$ is sampled with probability proportional to its frequency $f(w^{\prime})$. Using  unigram frequency for the noise distribution is common practice in \textit{noise contrastive estimation} \cite{dyer2014notes}.
    \item \textsc{UNIROOT}: $\mathbf{N}(w^{\prime}) = \frac{\sqrt{f(w^{\prime})}}{\sum_{w \in W} \sqrt{f(w)}}$ i.e a word $w^{\prime}$ is sampled with probability proportional to the square root of its frequency. Using such smoothed versions of the unigram frequency distribution has precedent in other NLP tasks. For instance, in \cite{goldberg2014word2vec}\footnote{Specifically, see the footnote concluding Page 2 in the paper}, the word2vec negative sampling uses $P(c) = \frac{{f_c}^{3/4}}{Z}$ to sample negative contexts, $c$, proportional to frequency, $f_c$. 
\end{enumerate}

\subsubsection{Choice of $\mathbf{B}$ architecture}
Since this is hitherto the first work to extend \textsc{LLR} method for NLP, we try to use a simple and standard architecture for $\mathbf{B}$. We use a left-to-right LSTM language model \cite{sundermeyer2012lstm} with a single layer. We vary the hidden state $\mathbf{B}_{h} \in \{64,128,256\}$. Note that $\mathbf{B}$ is non-class conditional - it does not use the labels in any way.

An additional point of consideration is that  $\mathbf{B}$ should not have a very large number of parameters, or have large time complexity at test time. Even in this regard, a \textit{LSTM} language model with a small state size is apt. We refer to approaches which use this architecture for $\mathbf{B}$ with +\textsc{BackLM}

\subsection{Generative Classifier} \label{subsec:Lgen}
Typical classification models estimate the conditional probability of the label given the input, i.e $P(y|x)$. An alternative paradigm  learns to estimate $P(x|y)$, additionally estimating $P(y)$ from the training set label ratios. Using Bayes rule, 
\begin{align*}
\argmax P(y|x) &= \argmax \frac{P(x|y) P(y)}{P(x)}  \\
               &= \argmax P(x|y) P(y) \\
\end{align*} %
Classifiers of this paradigm are called \textit{generative classifiers}, in contrast to the typical \textit{discriminative classifiers}.

\cite{yogatama2017generative} compare the two paradigms and found generative classifiers useful for  a) High sample efficiency b) Continual Learning c) Explicit Marginal Likelihood. The last point is particularly useful for us since we can use the explicit marginal likelihood from the classifier $P(x)$ as our $\eta$ function. Specifically, we use the $P(x) = \sum_{y \in Y} P(x|y)P(y)$  term which is directly available from a trained generative classifier. Hereon, we refer to this as $L_{gen}$.

\cite{yogatama2017generative} also propose a deep architecture for generative text classifiers that consists of a shared unidirectional LSTM across classes and a label embedding matrix. The respective label embedding is concatenated to the current hidden state and a final layer is then applied on this vector to give the distribution over the next word. The per-word cross-entropy loss serves as the loss function. We use a similar architecture as illustrated in Figure 1.

\section{Evaluation}
\label{sec:evaluation}
For the threshold dependent measures, we tune our $\eta()$ function on the validation set. We use the following metrics to measure \textit{OOD} detection performance:

\begin{itemize}
    \item \textbf{$FPR@k\%TPR$}: On picking a threshold such that the \textit{OOD} recall is  k\%, what fraction of the predicted \textit{OOD} points are \textit{ID}? FPR denotes False Positive Rate and TPR denotes True Positive Rate. Note that \textit{Positive} here refers to the \textit{OOD} class. We choose a high values of $k$, i.e $95$. Note that lower this value, the better is our \textit{OOD} Detection.
    \item \textbf{$AUROC$} : Measures the area under the Receiver Operating Characteristic, also known as the \textit{ROC} curve. Note that this curve is for the \textit{OOD} class. \cite{hendrycks2016baseline} first proposed using this. Higher the value, better is our \textit{OOD} Detection. This metric is threshold independent.
    \item \textbf{${AUPR}_{OOD}$ } : Area under the Precision Recall Curve is another threshold independent metric, based on the Precision-Recall Curve. Unlike $AUROC$, $AUPR$ is insensitive to class imbalance \cite{davis2006relationship}. The ${AUPR}_{OOD}$ and ${AUPR}_{ID}$ correspond to taking \textit{ID} and \textit{OOD} respectively as the positive class.
    %\item \textbf{$REC@95\%PR$} : 
\end{itemize}

\section{Datasets} \label{sec:data}
We use two datasets for our experiments. The first, \textsc{SNIPS}, is a widely used, publicly available dataset, and does not contain actual \textit{OOD} intents. The second, \textsc{ROSTD}, is a combination of a dataset released earlier \cite{schuster2018cross}, with new \textit{OOD} examples collected  by us. We briefly describe both of these in order. Table 1 also provides useful summary statistics about these datasets:

\subsection{SNIPS}
Released by \cite{coucke2018snips}, \textsc{SNIPs} consists of $\approx 15,000$ sentences spread through $7$ intent classes such as \textit{GetWeather}, \textit{RateBook} etc. As discussed previously, it does not explicitly include \textit{OOD} sentences. 

We follow the procedure described in \cite{lin2019deep} to synthetically create OOD examples. Intent classes covering atleast $K\%$ of the training points in combination are retained as $ID$. Examples from the remaining  classes are treated as \textit{OOD} and removed from the training set. In the validation and test sets,  examples from these classes are relabelled to the single class label \textit{OOD}. Besides not being genuinely \textit{OOD}, another issue with this dataset is that the validation and test splits are quite small in size at $700$ each.

In \S \ref{sec:experiments}, we report experiments on $K=75$ and $K=25$\footnote{See appendix for the experimental results with $K=25$ i.e \textsc{SNIPS},25\%.}, both of which ratios were used in \cite{lin2019deep}. We refer to these datasets as \textsc{SNIPS},75\% and \textsc{SNIPS},25\% respectively. Since multiple \textit{ID}-\textit{OOD} splits of the classes satisfying these ratios are possible, our results are averaged across 5 randomly chosen splits.  

\subsection{ROSTD}  \label{section:newDataset}

We release a dataset of $\approx 4590$ \textit{OOD} sentences . These sentences were curated to be explicitly \textit{OOD} with respect to the English split of the recently released dataset of intents from \cite{schuster2018cross} as the \textit{ID} dataset. This dataset contained $43,000$ intents from $13$ intent classes. We chose this dataset over \textit{SNIPs} owing to its considerably larger size ($\approx 2.3$ times larger). The sentences were authored by human annotators with the instructions as described in the subsection \textbf{Annotation Guidelines}.

\begin{table}
    %\centering
    \captionsetup{font=footnotesize}
    %\captionsetup{font=small}
    \scriptsize
    %\tiny
    \setlength{\tabcolsep}{4pt}
    \begin{tabular}{|l|l|l|}
        \hline \textbf{Category} &  \multicolumn{1}{|l|}{ \textbf{Example}} & \begin{tabular}{l} \textbf{\%}  \end{tabular}   \\ \hline
        
        \begin{tabular}{l} Overtly Powerful \\ Action \end{tabular}  & \begin{tabular}{l} 1. send Ameena \$ 25 from  Venmo account \\
        2. fix a pot of coffee
        \end{tabular} & 20.55  \\ \hline
        
        \begin{tabular}{l} Action \\ Memory \end{tabular} & \begin{tabular}{l} 1. What's the color of the paint I \\ bought off Amazon \\ 2. how much did I spend yesterday \end{tabular} & 12.24  \\ \hline
        
        \begin{tabular}{l} Declarative \\ Statement \end{tabular}   & \begin{tabular}{l} 1. I learned some good words. \\ 2. I always bookmark my favorite website \\ to go back in it anytime. \\  3. all Star Wars movie are great  \end{tabular} & 8.74  \\ \hline
        
        \begin{tabular}{l} Underspecified \\ Query \end{tabular}   & \begin{tabular}{l} 1. On what website can I order medication? \\ 2. how many jobs is having been lost \end{tabular} & 33.94  \\ \hline
        
        \begin{tabular}{l} Speculative \\ Question \end{tabular} & \begin{tabular}{l} 1. Can I do all of my Amazon shopping \\ through the app? \\ 2. when is the next episode of General Hospital \end{tabular} &  6.91  \\ \hline
                
        \begin{tabular}{l} Subjective \\ Question \end{tabular} & \begin{tabular}{l} 1. What color goes well  with navy blue? \\ 2. where can I learn something new every day? \end{tabular} &  27.99  \\ \hline
        
    \end{tabular}
    \caption{We manually classify each \textit{OOD} sentence in \textsc{ROSTD} into [1 or more] of 6 qualitative categories named self-explanatorily. More examples per category can be seen in Table 1 of the appendix.
    }
    \label{tab:categories}
\end{table}

\subsubsection{Annotation Guidelines} \label{subsubsec:annotationGuidelines}
We use human annotators to author intents which are explicity with respect to the English split of \cite{schuster2018cross}. The requirements and instructions for annotation were as follows:

\begin{enumerate}
    \item The \textit{OOD} utterances were authored by several distinct English-speaking annotators from Anglophone countries. 
    \item The annotators were asked to author sentences which were both grammatical and semantically sensible as English sentences. This was to prevent our \textit{OOD} data from becoming trivial by inclusion of ungrammatical sentences, gibberish and nonsensical sentences.
    \item The annotators were well informed of existing intent classes to prevent them from authoring intents . This was done by presenting the annotators with $5$ examples from the training split of each intent class, with the option to scroll for more through a dropdown. 
    \item After the first round of annotators had authored such intents, each intent was  post-annotated as in-domain vs out-of-domain by two fresh annotators who were not involved in the authoring stage.
    \item If both annotators agreed that the example was \textit{OOD}, it was retained. If both agreed it was \textit{ID}, it was discarded. In the event the two annotators disagreed, an additional, third annotator was asked to label the example and adjudicate the disagreement.
    \item During post-processing, we removed utterances which were shorter than three words. 
\end{enumerate}

%The data collection process outlined in Sebastian's paper is exactly what we did for OOD too. 
%1. We asked native English speakers to generate conversational data consisting of commands and questions that do not align with any of the domains.
%2.  Two annotators would label the utterances as out of domain or in-domain
% 3. If annotators disagreed, then a 3rd annotator would adjudicate the disagreement.

\subsubsection{Qualitative Analysis} \label{qualAnalysis}
We identify six  qualitative categories which might be making the sentences \textit{OOD}. We then manually assign each example into these categories. We summarize their distribution in Table \ref{tab:categories}. Note that since these categories are not mutually exhaustive, an example may get assigned multiple categories. 

\subsection{Coarsening Labels} \label{coarseningLabels}
The \textit{ID} examples from \cite{schuster2018cross}, which we also use as the \textit{ID} portion of \textsc{ROSTD} has hierarchical class labels [e.g $alarm/set\_alarm$,  $alarm/cancel\_alarm$ and $weather/find$]\footnote{See \cite{schuster2018cross} for full list}. Hence, \textsc{ROSTD} has a large number of classes (12), not all of which are equally distinct from each other. To ensure that our results are not specific only to settings with this kind of hierarchical label structure, we also experiment with retaining only the topmost or most \textit{``coarse"} label on each example. We refer to this variant with ``coarsened" labels as \textsc{ROSTD-COARSE}.

\begin{table}
    %\captionsetup{font=footnotesize}
    \captionsetup{font=small}
    \centering
    \footnotesize %Change back to tiny to cut size
    %\begin{center}
    %\tiny
    \begin{tabular}{|l|l|l| }
    \hline 
    \textbf{Statistic} & \textbf{\textsc{ROSTD}} & \textbf{\textsc{SNIPS}}   \\ \hline  \hline
    Train-ID & 30521 & 13084  \\ 
    Valid-ID & 4181 & 700 \\ 
    Test-ID  & 8621  & 700  \\ 
    Actual OOD & 4590 & None \\ \hline 
    Unique Word Types & {11.5K} & 11.4K \\ 
    Unique Bigrams &  {47.3K} & 36.3K  \\ 
    Unique Trigrams & {80.8K} & 52.2K \\ \hline
    Mean Utterance Length & {6.85} & {6.79} \\ 
    Number of \textit{ID} classes & {12/3 (Coarse)} & {7} \\  \hline
    %Dataset Source & \url{gameknot.com} \\ \hline
    \end{tabular}
    %\end{center}
    \caption{Dataset and Vocabulary Statistics contrasting \textsc{ROSTD} and \textsc{SNIPS}. Note that the \textit{ID} part of \textsc{ROSTD} comes from the English portion of the publicly available data from \cite{schuster2018cross} }
    \textbf{\label{table:stats}}
\end{table} %

%\subsubsection{Removing Artifacts}
%\todo[inline]{ Use a super-shallow classifier based on BOW and Length to flag examples easily classifeable as \textit{OOD}.} 

%\subsection{NEW-HOLDOUT}
%We show that our curated \textbf{OOD} points in combination with the \textit{ID} points from \cite{schuster2018cross} actually form a challenging benchmark using this as a control. The English part of the \cite{schuster2018cross} , which we use, has $30521+4181+8621$ intents from $12$ domains.

%Similar to how \textit{OOD} points are created in the case of \textbf{SNIPS} , we hold out a fraction of training \textit{ID} classes and present them to the model only at test-time. 

%\input{sections/data.tex}
 
%D_{Train}
\section{Experiments}
\label{sec:experiments}
%\vspace{-1em}
\begin{table*}[t!]
%\begin{center}
\captionsetup{font=small}
\centering
\footnotesize
\begin{tabular}{l l l  l l  l}
\hline %\toprule
\bf Dataset & \bf Model & \bf $F_{1} \uparrow$  &  $FPR@95\%TPR \downarrow$ & AUROC $\uparrow$ & \bf{$AUPR_{OOD} \uparrow$}      \\ 
\hline %\midrule
\multirow{3}{*}{\textsc{ROSTD}} & \textsc{MSP} & 54.22 $\pm$ 4.01  & 100.00 $\pm$ 0.00 & 70.75 $\pm$ 3.70 &  55.68 $\pm$ 6.36   \\
& \textsc{MSP,{$\tau=1e^{3}$}} & 55.45 $\pm$ 4.19  & 60.48 $\pm$ 3.17 & 76.94 $\pm$ 4.01 & 59.64 $\pm$ 6.49   \\
%& \textsc{PPT} & 41.77 $\pm$ x & 100.0 $\pm$ x & 100.0 $\pm$ x & 59.03 $\pm$ x &  60.90 $\pm$ x & 88.50 $\pm$ x  \\
& \textsc{$-KL(P|\mathbf{U})$} & 55.31 $\pm$ 3.89  & 60.15 $\pm$ 3.11 & 76.86 $\pm$ 3.85 & 59.54 $\pm$ 6.10   \\
& \textsc{$-KL(P|\mathbf{R})$} & 83.24 $\pm$ 2.78  & 21.31 $\pm$ 8.38 & 95.78 $\pm$ 1.30 & 90.32 $\pm$ 1.97   \\
& \textsc{LOF} & 64.46 $\pm$ 2.57  & 42.49 $\pm$ 3.49 & 81.39 $\pm$ 2.38 &  46.89 $\pm$ 3.50   \\
& \textsc{LOF+LMCL} & 85.97 $\pm$ 2.00  & 15.03 $\pm$ 5.42 & 95.60 $\pm$ 0.75 &  82.71 $\pm$ 9.17    \\
& {\textsc{$\mathbf{L}_{simple}$}} & 81.38 $\pm$ 0.19  & 18.92 $\pm$ 0.56 &  95.42 $\pm$ 0.11 & 87.38 $\pm$ 0.41  \\
& {\textsc{$\mathbf{L}_{simple}$}+\textsc{BackLM}+\textsc{UNIFORM}} & 85.25 $\pm$ 0.72  & 36.65 $\pm$ 6.87 &  94.71 $\pm$ 0.49 & 91.10 $\pm$ 0.63  \\
& {\textsc{$\mathbf{L}_{simple}$}+\textsc{BackLM}+\textsc{UNIGRAM}} & 82.27 $\pm$ 0.74  & 42.16 $\pm$ 3.62 &  93.62 $\pm$ 0.43 & 89.30 $\pm$ 0.50  \\
& {\textsc{$\mathbf{L}_{simple}$}+\textsc{BackLM}+\textsc{UNIROOT}} & 87.42 $\pm$ 0.45  & 20.10 $\pm$ 5.25 &  96.35 $\pm$ 0.41 & 93.44 $\pm$ 0.37  \\
& {\textsc{$\mathbf{L}_{gen}$}} & 86.25 $\pm$ 0.71 & 10.86 $\pm$ 1.08 & 97.42 $\pm$ 0.28 & 92.30 $\pm$ 0.99   \\
& {\textsc{$\mathbf{L}_{gen}$}+\textsc{BackLM}+\textsc{UNIFORM}} & 89.60 $\pm$ 0.56   & 13.71 $\pm$ 5.64 &  97.67 $\pm$ 0.35 & 95.49 $\pm$ 0.42  \\
& {\textsc{$\mathbf{L}_{gen}$}+\textsc{BackLM}+\textsc{UNIGRAM}} & \bf 91.35 $\pm$ 2.62  & 10.55 $\pm$ 4.11 &  97.87 $\pm$ 0.49 & 95.86 $\pm$ 0.68  \\
& {\textsc{$\mathbf{L}_{gen}$}+\textsc{BackLM}+\textsc{UNIROOT}} &  91.17 $\pm$ 0.32  & \bf 7.41 $\pm$ 1.88 &  \bf 98.22 $\pm$ 0.26 & \bf 96.47 $\pm$ 0.29  \\
%& {\textsc{$\mathbf{L}_{gen}$}+\textsc{BackSelf}+\textsc{UNIROOT}} & 90.04 $\pm$ 0.35 & 01.33 $\pm$ 0.17 & 7.60 $\pm$ 0.58 & 97.87 $\pm$ 0.22 & 95.86 $\pm$ 0.18 & 98.95 $\pm$ 0.16 \\
\hline %\midrule
\multirow{3}{*}{\textsc{ROSTD-COARSE}} & \textsc{MSP} & 59.99 $\pm$ 19.01  & 26.00 $\pm$ 34.32 & 71.63 $\pm$ 15.55 &  64.32 $\pm$ 19.36   \\
& \textsc{MSP,{$\tau=1e^{3}$}} & 64.62 $\pm$ 15.31  & 64.46 $\pm$ 9.84 & 78.39 $\pm$ 11.92 & 66.89 $\pm$ 11.76   \\
%& \textsc{PPT} & 41.77 $\pm$ x & 100.0 $\pm$ x & 100.0 $\pm$ x & 59.03 $\pm$ x &  60.90 $\pm$ x & 88.50 $\pm$ x  \\
& \textsc{$-KL(P|\mathbf{U})$} & 65.36 $\pm$ 15.49  & 65.39 $\pm$ 4.84 & 79.05 $\pm$ 11.40 & 67.79 $\pm$ 19.43   \\
& \textsc{$-KL(P|\mathbf{R})$} & 81.56 $\pm$ 8.51  & 17.78 $\pm$ 15.70 & 93.47 $\pm$ 6.25 & 87.49 $\pm$ 8.94   \\
& \textsc{LOF} & 62.39 $\pm$ 9.01  & 46.55 $\pm$ 17.56 & 78.07 $\pm$ 12.23 &  45.80 $\pm$ 12.21   \\
& \textsc{LOF+LMCL} & 84.28 $\pm$ 3.44  & 15.24 $\pm$ 4.70 & 95.19 $\pm$ 1.03 &  76.63 $\pm$ 2.53    \\
& {\textsc{$\mathbf{L}_{simple}$}} & 80.48 $\pm$ 0.27  & 20.78 $\pm$ 0.71 &  95.2 $\pm$ 0.07 & 86.87 $\pm$ 0.13  \\
& {\textsc{$\mathbf{L}_{simple}$}+\textsc{BackLM}+\textsc{UNIFORM}} & 85.97 $\pm$ 0.65  & 30.65 $\pm$ 4.51 & 95.27 $\pm$ 0.47 & 91.98 $\pm$ 0.61  \\
& {\textsc{$\mathbf{L}_{simple}$}+\textsc{BackLM}+\textsc{UNIGRAM}} & 84.46 $\pm$ 0.62  & 31.79 $\pm$ 3.04 &  94.93 $\pm$ 0.32 & 91.22 $\pm$ 0.50  \\
& {\textsc{$\mathbf{L}_{simple}$}+\textsc{BackLM}+\textsc{UNIROOT}} & 88.25 $\pm$ 0.50  & 16.35 $\pm$ 1.32 &  96.82 $\pm$ 0.12 & 94.10 $\pm$ 0.20  \\
& {\textsc{$\mathbf{L}_{gen}$}} & 86.67 $\pm$ 0.34  & 9.88  $\pm$ 0.44 & 97.58 $\pm$ 0.08 & 92.74 $\pm$ 0.29   \\
& {\textsc{$\mathbf{L}_{gen}$}+\textsc{BackLM}+\textsc{UNIFORM}} & 89.32 $\pm$ 0.30   & 8.04 $\pm$ 0.69 &  97.83 $\pm$ 0.15 & 95.27 $\pm$ 0.32  \\
& {\textsc{$\mathbf{L}_{gen}$}+\textsc{BackLM}+\textsc{UNIGRAM}} & 90.05 $\pm$ 0.73    & \bf 6.69 $\pm$ 0.82 & 98.16 $\pm$ 00.15  & 95.61  $\pm$ 0.50  \\
& {\textsc{$\mathbf{L}_{gen}$}+\textsc{BackLM}+\textsc{UNIROOT}} & \bf 90.14 $\pm$ 0.39  & 6.78 $\pm$ 0.60 & \bf 98.30 $\pm$ 0.09 & \bf 95.96 $\pm$ 00.37  \\
%& {\textsc{$\mathbf{L}_{gen}$}+\textsc{BackSelf}+\textsc{UNIROOT}} & 90.04 $\pm$ 0.35 & 01.33 $\pm$ 0.17 & 7.60 $\pm$ 0.58 & 97.87 $\pm$ 0.22 & 95.86 $\pm$ 0.18 & 98.95 $\pm$ 0.16 \\
%  \multirow{3}{*}{\textsc{Large}} & \textsc{MSP} & 23.11  & 35.53  & 68.48 & 79.71  & 14.5  & 98.92   \\
%  & \textsc{MSP, $\tau=1e^{3}$} & 22.08  & 32.57  & 63.10  & 80.76  & 14.27  & 99   \\
% & \textsc{$KLD(P|\mathbf{U})$} & 22.12  & 32.61 & 63.06  & 80.73  & 14.03  & 99   \\
% & \textsc{$KLD(P|\mathbf{R})$} & 25.05 & \textbf{30.25} & 57.12  & 82.54  & 16.58 & 99.14   \\
% & {\textsc{$\mathbf{L}_{simple}$}} & 20.91  & 20.91  & \textbf{50.09}  & \textbf{82.74}  & 13.66  & \textbf{99.16}   \\
% & {\textsc{$\mathbf{L}_{simple}$}+\textsc{BackLM}} & \textbf{31.79}  & 39.63  & 71.75  &  80.26  & 22.94  & 98.88  \\
% & {\textsc{$\mathbf{L}_{gen}$}} & 19.81  & 30.04 & 52.71 &  81.61 &  12.53 & 99.14  \\
% & {\textsc{$\mathbf{L}_{gen}$}+\textsc{BackLM}} & 29.14  & 39.13  & 70.99 & 79.66  & 22.23 & 98.88 \\
% & {\textsc{$\mathbf{L}_{gen}$}+\textsc{BackSelf}} & 30.37  & 44.42  & 72.18 &  78.90  & \textbf{23.69}  & 98.81  \\
\hline %\midrule
  \multirow{3}{*}{\textsc{SNIPS, 75\%}}  & \textsc{MSP} & 81.58 $\pm$ 7.68  & \bf 16.68 $\pm$ 18.06 & 93.51 $\pm$ 4.49 & 85.03 $\pm$ 6.19   \\
  & \textsc{MSP,{$\tau=1e^{3}$}} & 83.94 $\pm$ 6.82  & 31.32 $\pm$ 30.25 & 94.30 $\pm$ 4.50 & 88.44 $\pm$ 5.72   \\
& \textsc{$-KL(P|\mathbf{R}) \approx -KL(P|\mathbf{U})$} & 84.23 $\pm$ 7.22  & 29.28 $\pm$ 27.04 & 94.51 $\pm$ 4.38 & 88.71 $\pm$ 6.15   \\
& \textsc{LOF} & 66.07 $\pm$ 8.82  & 49.56 $\pm$ 13.49 & 79.65 $\pm$ 7.81 & 51.69 $\pm$ 13.08   \\
& \textsc{LOF+LMCL} & 76.24 $\pm$ 9.34  & 42.27 $\pm$ 20.64 & 90.37 $\pm$ 6.54 & 77.81 $\pm$ 10.53   \\
 & {\textsc{$\mathbf{L}_{simple}$}} & 63.51 $\pm$ 6.33  & 54.56 $\pm$ 12.13 & 81.72 $\pm$ 5.90 & 62.12 $\pm$ 13.28   \\
 & {\textsc{$\mathbf{L}_{simple}$}+\textsc{BackLM}+\textsc{UNIFORM}} & 74.74 $\pm$ 3.25  & 44.60 $\pm$ 12.01  & 90.02 $\pm$ 2.24 & 80.08 $\pm$ 3.31  \\
  & {\textsc{$\mathbf{L}_{simple}$}+\textsc{BackLM}+\textsc{UNIGRAM}} & 81.19 $\pm$ 3.53  & 27.00 $\pm$ 8.71 &  93.97 $\pm$ 1.85 & 87.57 $\pm$ 3.38  \\
   & {\textsc{$\mathbf{L}_{simple}$}+\textsc{BackLM}+\textsc{UNIROOT}} & 78.75 $\pm$ 3.25  & 35.24 $\pm$ 11.08 &  92.66 $\pm$ 1.89 & 84.84 $\pm$ 3.36  \\
 & {\textsc{$\mathbf{L}_{gen}$}} & 67.31 $\pm$ 7.06  & 44.60 $\pm$ 18.53 &  85.17 $\pm$ 7.18 & 68.11 $\pm$ 14.29  \\
 & {\textsc{$\mathbf{L}_{gen}$}+\textsc{BackLM}+\textsc{UNIFORM}} & 78.37 $\pm$ 6.60 & 29.28 $\pm$ 4.23 &  92.35 $\pm$ 2.77 & 82.84 $\pm$ 7.33  \\
 & {\textsc{$\mathbf{L}_{gen}$}+\textsc{BackLM}+\textsc{UNIGRAM}} & \bf 85.47 $\pm$ 6.90  & 18.48 $\pm$ 11.26 & \bf 95.79 $\pm$ 2.67 & \bf 90.98 $\pm$ 6.73  \\
 & {\textsc{$\mathbf{L}_{gen}$}+\textsc{BackLM}+\textsc{UNIROOT}} & 81.91 $\pm$ 6.83  & 22.24 $\pm$ 6.26 &  94.15 $\pm$ 2.59 & 86.60 $\pm$ 7.03  \\
% & {\textsc{$\mathbf{L}_{gen}$}+\textsc{BackSelf}} & \textbf{x} $\pm$ x & \textbf{x} $\pm$ x & \textbf{x} $\pm$ x &  x $\pm$ x & \textbf{x} $\pm$ x & x $\pm$ x \\
%\hline
%\multirow{3}{*}{\textsc{\textsc{SICK}}} & \textsc{BERTbase} & xx $\pm$ xx  & 53.8 $\pm$ 0.0 & 48.05 $\pm$ 0.0  &   &  xx $\pm$ xx & xx $\pm$ xx \\
%& \textsc{AUX-$L_1$} & xx $\pm$ xx & xx $\pm$ xx & xx $\pm$ xx   &  xx $\pm$ xx & xx $\pm$ xx & xx $\pm$ xx \\
%& \textsc{HDW} & xx $\pm$ xx & 55.33 $\pm$ 0.0 & 48.61 $\pm$ 0.0  &  xx $\pm$ xx & xx $\pm$ xx & xx $\pm$ xx   \\
%\textsc{SpaceStateReg} & xx & xx  \\
\hline
%\bottomrule

\end{tabular}
\normalsize
%\end{center}
\caption{Performance of the baseline methods and our proposed models on \textsc{ROSTD}, \textsc{ROSTD-COARSE} and \textsc{SNIPS}. $\downarrow$ ($\uparrow$)  indicates lower (higher) is better. We can see that the $\mathbf{L}_{gen}+$\textsc{BackLM}+\textit{$<$Noise$>$} (where \textit{$<$Noise$>$} is one of three noising schemes) approaches outdo their non \textsc{LLR} counterparts on most measures. For \textsc{SNIPS}, $-KL(P|\mathbf{R})\approx -KL(P|\mathbf{U})$ since the training set is almost evenly distributed between the \textit{ID} classes. We can also observe that the differences in performance between different approaches are much more observable on \textsc{ROSTD} as compared to \textsc{SNIPS}.}
\label{tab:mainres}
%\vspace{-0.4cm}
\end{table*}

We compile the results of all our experiments in Table \ref{tab:mainres}. 

\subsection{Implementation}
All experiments are averaged across 5 seeds. We use Pytorch 1.0 \cite{paszke2017automatic} to implement models\footnote{Code available at \url{github.com/vgtomahawk/LR_GC_OOD}}. 

The checkpoint with highest validation F1 on the \textit{ID} subset of the validation set is chosen as the final checkpoint for computing the other \textit{OOD} evaluation metrics. For the label-agnostic approaches ($L_{simple}$), the checkpoint with lowest validation perplexity is chosen. For the +\textsc{BackLM} approaches, we use $p_{noise}=0.5$. We also experimented with $p_{noise} \in \{0.1,0.3,0.7\}$, but find $0.5$ works best.

\subsubsection{Base Classifier Architectures}
For the discriminative classifier, we use a bidirectional LSTM [1-layer] with embedding size 100, projection layer of 100$\times$300 (to project up embeddings), hidden size $300$ and embeddings initialized with Glove (\emph{glove.6B.100D}) \cite{pennington2014glove}. Generative classifier approaches have similar architecture except that they are unidirectional and have additional label embeddings of dimension $20$

\subsubsection{LOF implementation}
We use the scikit-learn 0.21.2 implementation\footnote{\url{https://scikit-learn.org/stable/modules/generated/sklearn.neighbors.LocalOutlierFactor.html}} \cite{pedregosa2011scikit} of \textit{LOF}. We fix the number of nearest neighbors to $20$ but tune the contamination rate as a hyperparameter. We also corroborated over email correspondence with the authors of \cite{lin2019deep} that they had used a similar hyperparameter setting for \textsc{LOF}.

%TODO LIST
%\begin{enumerate}
    %\item \todo[inline]{Graph showing scatter plot of ID vs OOD dev points using $L_{gen}$ vs $L_{gen}$+\textsc{BackLM}}
    %\item \todo[inline,color=green]{Graph showing scatter plot of ID vs OOD dev points using $L_{simple}$ vs $L_{gen}$}
    %\item \todo[inline,color=yellow]{Scatter plot showing backgrounds improve for multiple base architectures}
    %\item \todo[inline,color=magenta]{Best methods performance on different OOD vs ID ratios}
    %\item \todo[inline,color=orange]{Background performance sensitive to different noise levels}
    %\item \todo[inline,color=blue]{Comparison of \textsc{RELEASE} actual and synthetic}
    % \item \todo[inline,color=blue]{Fill up the table with results on multiple seeds} DONE
%\end{enumerate}

\begin{figure*}[!htb] %[ht]
    \captionsetup{font=small}
    \centering
    %\begin{subfigure}{0.20\textwidth}
    %\captionsetup{font=footnotesize}
    %\fbox{\includegraphics[width=\textwidth]{Project/figures/related/GameImages/test0.png}}
    %\caption{Test 0}
    %\end{subfigure}
    %\begin{subfigure}{0.20\textwidth}
    %\captionsetup{font=footnotesize}
    %\fbox{\includegraphics[width=\textwidth]{Project/figures/related/GameImages/test1.png}}
    %\caption{Test 1}
    %\end{subfigure}
    %\begin{subfigure}{0.20\textwidth}
    %\captionsetup{font=footnotesize}
    %\fbox{\includegraphics[width=\textwidth]{Project/figures/related/GameImages/test2.png}}
    %\caption{Test 2}
    %\end{subfigure}
   %\begin{subfigure}{0.20\textwidth}
    %\captionsetup{font=footnotesize}
    %\fbox{\includegraphics[width=\textwidth]{Project/figures/related/GameImages/test3.png}}
    %\caption{Test 3}
    %\end{subfigure}
    %\\
     \begin{subfigure}{0.32\textwidth}
    \captionsetup{font=footnotesize}
    \includegraphics[width=\textwidth]{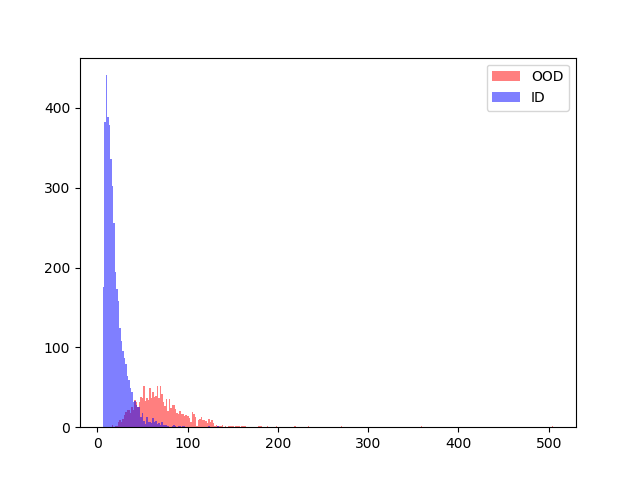}
    \caption{\textsc{${\mathbf{L}}_{gen}$}}
    \end{subfigure}
    \begin{subfigure}{0.32\textwidth}
    \captionsetup{font=footnotesize}
    \includegraphics[width=\textwidth]{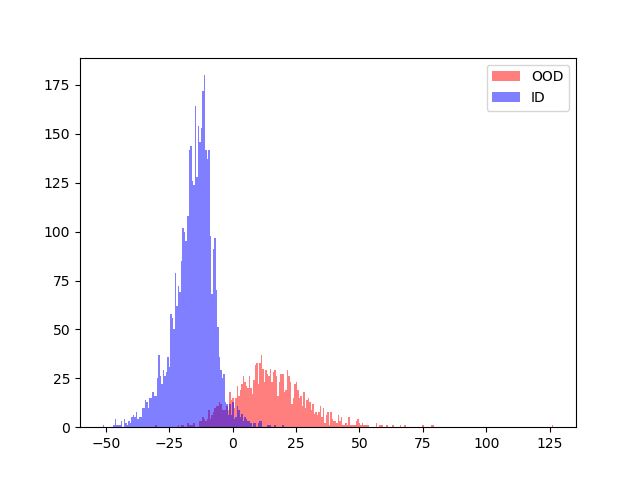}
    \caption{\textsc{${\mathbf{L}}_{gen}$+BackLM+UNIROOT}}
    \end{subfigure}

    \caption{Effect of \textsc{BackLM+UNIROOT}. In the right graph, we can see that the \textit{OOD} has shifted considerably to the right then before and overlaps less with the \textit{ID} set }
    \label{fig:AllValidationBPCCurves}
    %\vspace{-4mm}
\end{figure*} %

\subsection{Observations}
 From Table \ref{tab:mainres}, we see that $\mathbf{L}_{gen}$ outperforms uncertainty based and nearest neighbour approaches by a reasonable margin on both datasets. It is also significantly better than the language model likelihood based $\mathbf{L}_{simple}$. This validates our hypothesis that generative classifiers effectively combine the benefits of likelihood-based and uncertainty-based approaches.  

Furthermore, \textsc{LLR} based approaches always outperform the respective likelihood-only approach, whether $\mathbf{L}_{simple}$ or $\mathbf{L}_{gen}$. Amongst different noising methods, the performance improvement is typically largest using the  \textsc{UNIROOT} approach we proposed. For instance,  on \textsc{ROSTD},
\begin{equation*}
\scriptsize
\mathbf{L}_{gen}+\textsc{BackLM}+\textsc{Uniroot} > \mathbf{L}_{gen}+\textsc{BackLM}+\textsc{Uniform}
\normalsize
\end{equation*} %
\begin{equation*}
\scriptsize
 \mathbf{L}_{gen}+\textsc{BackLM}+\textsc{Uniform} >
 \mathbf{L}_{gen}
\normalsize
\end{equation*} %
\begin{equation*}
\scriptsize
\mathbf{L}_{simple}+\textsc{BackLM}+\textsc{Uniroot} > \mathbf{L}_{simple}+\textsc{BackLM}+\textsc{Uniform}
\normalsize
\end{equation*} %
\begin{equation*}
\scriptsize
 \mathbf{L}_{simple}+\textsc{BackLM}+\textsc{Uniform} >
 \mathbf{L}_{simple}
\normalsize
\end{equation*} %
A clear advantage of \textsc{ROSTD} which is clear from the experiments is that differences in performance between the various methods are much more pronounced when tested on it, as compared to \textsc{SNIPS}. On \textsc{SNIPS}, the simple \textsc{MSP, $\tau=1e^3$} baseline is itself able to reach $\approx$ 80-90\% of the best performing approach on most metrics.

\section{Conclusion}
\label{sec:conclusions}
To the best of our knowledge, we are hitherto the first work to use an approach based on generative text classifiers for \textit{OOD} detection. Our experiments show that this approach can outperform existing paradigms significantly on multiple datasets.

Furthermore, we are the first to flesh out ways to use likelihood ratio based approaches first formalized by \cite{ren2019likelihood} for \textit{OOD} detection in NLP. The original work had tested these approaches only for DNA sequences which have radically smaller vocabulary than NL sentences. We propose \textsc{UNIROOT}, a new way of noising inputs which works better for NL. Our method improves two different likelihood based approaches on multiple datasets. 

Lastly, we curate and plan to publicly release \textsc{ROSTD}, a novel dataset of \textit{OOD} intents w.r.t the intents in \cite{schuster2018cross}. We hope \textsc{ROSTD} fosters further research and serves as a useful benchmark for \textit{OOD} Detection%, for both intents in particular and natural language in general%\input{sections/conclusions.tex}
\section{Acknowledgements}
We thank Tony Lin and co-authors for promptly answering several questions about their paper, and Sachin Kumar for valuable discussion on methods. We also thank Hiroaki Hayashi and 3 anonymous reviewers for valuable comments.
%\newpage
\bibliography{AAAI-GangalV.4206}
\bibliographystyle{AAAI-GangalV.4206}

\end{document}